# Towards a Job Title Classification System

Faizan Javed, Matt McNair, Ferosh Jacob, Meng Zhao
Classification R&D,
CareerBuilder.com,
5550 Peachtree Parkway, Greater Atlanta Area, GA 30092, USA
{Faizan.Javed, Matt.McNair, Ferosh.Jacob, Meng.Zhao}@CareerBuilder.com

## ABSTRACT
Document classification for text, images and other applicable entities has long been a focus of research in academia and also finds application in many industrial settings. Amidst a plethora of approaches to solve such problems, machine-learning techniques have found success in a variety of scenarios. In this paper we discuss the design of a machine learning-based semi-supervised job title classification system for the online job recruitment domain currently in production at CareerBuilder.com and propose enhancements to it. The system leverages a varied collection of classification as well clustering algorithms. These algorithms are encompassed in an architecture that facilitates leveraging existing off-the-shelf machine learning tools and techniques while keeping into consideration the challenges of constructing a scalable classification system for a large taxonomy of categories. As a continuously evolving system that is still under development we first discuss the existing semi-supervised classification system which is composed of both clustering and classification components in a proximity-based classifier setup and results of which are already used across numerous products at CareerBuilder. We then elucidate our long-term goals for job title classification and propose enhancements to the existing system in the form of a two-stage coarse and fine level classifier augmentation to construct a cascade of hierarchical *vertical classifiers*. Preliminary results are presented using experimental evaluation on real world industrial data.

## Categories and Subject Descriptors
I.5.2 [**Pattern Recognition**]: Design Methodology – *classifier design and evaluation.*
I.5.4 [**Pattern Recognition**]: Applications – *text processing.*

## General Terms
Algorithms, Experimentation, Performance

## Keywords
Semi-supervised Classification, Clustering, Text Classification

## 1. INTRODUCTION
Document classification is a widely studied problem in data mining, information retrieval and machine learning. While document classification is of great academic research interest, it also finds many applications in industrial settings. In this paper our focus is on text classification of job titles to a predefined set of occupation categories. Machine learning techniques have been successfully applied to text classification [1]. The learning approach to document classification entails assigning documents (which may be images, text, or other entities) a label of a predefined class or category to create a set of training data. This training dataset is used to learn a model that is then utilized to assign a class to new documents. With the proliferation of *Big Data* in the e-commerce and web domains, there has been an explosion in the amount of data that needs to be processed in a fast, efficient and scalable manner. Such large-scale e-commerce systems need to classify millions of items into thousands of categories to facilitate item catalog categorization to improve the customer experience of searching and browsing leading to more purchases and sales transactions.

In the online job recruitment domain, we face a similar problem of categorizing job titles. CareerBuilder.com has hundreds of millions of job postings, resumes, and job applications. The classification of these large datasets into a predefined set of job titles benefits two important applications. Firstly, the classifications are used to improve our search and recommendation products by making it easier for job seekers and employers to find each other. This facilitates CareerBuilder.com's goal of empowering employment and helping job keepers find the jobs and training they need. Secondly, good classification enables data to be aggregated, which in turn enables analysis. This is critical to our labor market analytics products (e.g. Supply and Demand), which give valuable insights to employers as they design their recruitment strategies.

In the past, we have classified our content into a standardized hierarchy of job titles commonly used in our industry known as O*NET[1]. O*NET is an extension of the Standard Occupational Classification (SOC)[2] system developed by the U.S. Bureau of Labor Statistics. The O*NET categorization system is a four level hierarchy with 23 major top level groups, 97 minor groups in the second level, 46 broad occupations in the third level, and 1,110 leaf level occupations. For example, the *Software Developers, Application* leaf level occupation is designated the O*NET code 15-1132.00 which corresponds to major group 15, broad occupation group 1132, minor group 1130 (*Software Developers & Programmers*), and the final two digit O*NET code 00 which in this instance is the same as minor group 1130 (i.e., *Software Developers, Application*). We have found that this taxonomy is not specific enough for our needs and is not updated frequently enough (at the time of this writing the last update was in 2010). For example, there are two SOC codes that represent software development. These two categories encompass all general and niche/emerging software engineering job titles such as *Hadoop Engineer, .Net Developer, Machine Learning Engineer* and *Java Developer* amongst others. The result is that our search and recommendation products are not able to refine the search results to the degree required to provide the best possible experience for our end users. This lack of granularity is the primary motivation for our work in this area.

---

[1] http://www.onetcenter.org/taxonomy/2010/list.html

[2] http://www.bls.gov/soc/major_groups.htm

Several techniques have been used for text classification such as Decision Trees [2], Naïve Bayes [3], K-Nearest-Neighbor (k-NN) [4] and Support Vector Machines (SVM) [5]. Popular machine learning toolkits and frameworks such as WEKA [6], Scikit [7] and LIBLINEAR [8] incorporate these algorithms for practical use. The large number of data instances and features along with increased computational requirements for complex learning models and (near) real-time inference constraints have resulted in new efforts that adapt machine-learning techniques to scale out/distributed and scale up architectures [9]. Some examples of large-scale distributed machine learning frameworks are Apache Mahout[3], Jubatus[4], GraphLab [10] and MLbase [11].

In this paper we describe a semi-supervised job title classification system with the pre-existing O*NET-SOC (henceforth stated as O*NET) hierarchy as target classes. Our existing classification approach is a flat classification system where the focus is on discovering and classifying job title labels. The system is composed of a clustering component that discovers job titles from a jobs dataset that are then used by an instance based, multi-label k-NN classifier that classifies jobs to these cluster labels. Note that currently these cluster labels are not O*NET categories, but job title strings. This is a *big bang* approach to classification because a single classifier is trained on a dataset that represents the entire target hierarchy. As part of our work towards categorizing jobs to O*NET categories, we enhance the system by decomposing it into hierarchical coarse and fine level classifiers where a fine level classifier is created for every SOC major group resulting in a *vertical classifier*. Our long-term goal is to have two classification pipelines: one which classifies job titles to O*NET categories, and the other which is the existing job title label classification. Both will share the same coarse-level classifier. Since data for text classification is usually sparse and high dimensional, these characteristics make SVMs and their ability to generalize well in high dimensional spaces and speed of classification an especially good fit for text classification problems. We use an SVM classifier at the coarse level and the current multi-label k-NN classifier at the fine level. We show that using this approach of treating classes separately alleviates processing time, enhances scalability, and results in more accurate job title classification with a finer granularity in the set of titles discovered.

## 2. RELATED WORK
Concept hierarchies and taxonomies exist in many diverse domains such as web portals and e-commerce to library systems and medical text. Some examples are Wikipedia, Amazon.com, Ebay.com, the U.S. Library of Congress Classification and the International Classification of Diseases amongst others. Extensive work has been done in leveraging pre-existing hierarchies for classification and also making use of clustering techniques to overcome the scarcity of labeled data for training classifiers. Dumais and Chen [12] demonstrated the first use of SVMs for classifying a large heterogeneous collection of web content. They use SVMs as both first and second level classifiers with a threshold score to be exceeded at the top level before a comparison can be made at the second level. In [13] Xue et al propose a *Deep Classification* method which consists of two stages: the first stage uses a category search to obtain a set of candidates for the test document which are then used to prune the hierarchy and train a classifier. The main problem with this approach is the computational requirement of training a classifier for every instance that needs to be classified. Ruiz and Srinivasan apply an "expert and gates" strategy in concert with a hierarchical array of neural networks to the text categorization problem of medical records in [14]. Slonim and Tishby [15] demonstrate the power of using word clusters to improve text classification. They apply the information bottleneck principle to create word-clusters and reduce the feature space dimensionality. Malik and Kender [16] introduce CPHC, a semi-supervised classification algorithm that utilizes a pattern-based cluster hierarchy to directly classify instances, which alleviates the need to train a classifier. In [17] Malik proposes a two-stage hierarchical SVM text classification system that mitigates the impact of compounding errors inherent in hierarchical SVMs by using $k^{th}$-level hierarchy flattening where k is user-defined. Crammer et al [18] use a cascading system of a k-best online learning algorithm and a rule-based and automatic coding policy for a multi-class, multi-label system to assign ICD-9-CM clinical codes to free text radiology reports. The output of the rule-based system is used as features for the online learning algorithm that improves its classification accuracy overtime. In [19], Lojo et al use k-NN and SVM to assign CIE-9MC codes to medical reports.

In terms of application domain and the approach taken, our work is most similar to LinkedIn's job title classification system [20] and Ebay's work on large-scale item categorization [21]. Similar to LinkedIn our focus is on job title classification but we use a hierarchical two-stage approach to classification while LinkedIn utilizes a heavily manual phrase-based classification system dependent on the near-sufficiency property of short-text and a heavy reliance on crowd-sourced labeling of training samples. Similar to Ebay we use a hierarchical approach of cascading classifiers such as k-NN and SVM for classification but our application domains differ. Also unlike Ebay we use SVM as a coarse level classifier and k-NN as a fine level classifier. In the next section we explain the design of our system.

## 3. JOB TITLE CLASSIFICATION
### 3.1 Proximity-based Classification
The existing job title classification system at CareerBuilder.com is a semi-supervised approach using a multi-class, multi-label proximity-based classifier. Proximity-based classifiers [22] use the observation that documents that belong to the same class are close to one another based on a distance metric such as the cosine distance. Such classifiers may find the k-nearest neighbors of a test instance and report the majority class, or to improve efficiency pre-process the data in clusters, define a meta-document for a cluster, and then find the k-nearest neighbors.

The first phase of our classification system uses a clustering algorithm to aggregate the training dataset into clusters. We use Lingo3G, a proprietary clustering library based on the Lingo clustering algorithm [23]. To balance both simplicity and accuracy, Lingo features a novel combination of both numerical and phrase-based clustering methods by identifying lexically meaningful and semantically distinct cluster labels before fulfilling cluster assignments. More specifically, the algorithm first conducts singular value decomposition (SVD) on the tf-idf [24] term-document matrix. Left-singular vectors are employed to identify cluster labels and singular values are used to determine a practical amount of clusters. Documents are then assigned to each clusters based on cosine distance over a user-defined threshold to quantify resemblance. The Lingo algorithm has a lower misclassification rate than other popular methods but consumes more memory due to relatively frequent and high dimensional matrix operations [25] especially for large datasets. For the existing system at Careerbuilder.com, we crawled the web to get 500k jobs with a job title minimum frequency threshold of four. These jobs were pre-processed by removing extraneous markup characters, stop-words and noise. The clustering process took around half an hour on a dual core i7 2Ghz Macbook Air with 8GB of RAM resulting in 1800 job title clusters ranging from *Nurse Assistant* to *Hadoop Engineer*. We note that the current O*NET taxonomy does not have an entry for emerging titles such as *Hadoop Engineer* and our approach which relies on a constantly evolving jobs dataset allows us to capture such niche titles. The k-NN classifier then uses these clusters as meta-documents for classification. While k-NN is considered a slow classifier due to its instance time classification properties, our classifier uses Lucene[5], an open-source, industrial strength search engine library that results in classification response time of less than 100ms. More specifically we use the Lucene *morelikethis* class to construct classification queries and use cosine similarity for document vector similarity. We used a combination of unigram and bigram terms to construct the queries to because there are both single word and multi-word job titles. We fetch the term vector for every document that is indexed, create a term frequency (tf) map according to a minimum frequency threshold, compute the inverse document frequency (idf) and output a list of titles that are a product of the tf-idf score. The classifier is currently deployed as an internal API and is also used by other teams for their products.

---

[3] http://mahout.apache.org/

[4] http://jubat.us/en/

## 3.2 Vertical Classifiers

The drawbacks of the current system are: 1) its current architecture is not suitable for achieving our long-term goal of an automated O*NET classification system, and 2) it does not scale with more training data which is what we need to obtain a more wide ranging list of job titles which go beyond the O*NET taxonomy. To this end we discuss an extension in the form of incorporating an SVM classifier as the first phase in pipeline. This phase involves classifying jobs to the appropriate top-level two-digit SOC groups which are then used in the existing k-NN pipeline resulting in *vertical classifiers* or k-NN classifiers for each of the 23 top level SOC codes. SVMs achieve state of the art performance on text classification: they do not over fit the data and are robust on sparse and high-dimensional data [5]. We chose SVM as the coarse level classifier because in the cascade of classifiers it is very important to get the top-level classification correct. SVMs are usually binary classifiers that aim to partition the data space with linear or non-linear delineations between the classes. We use LIBLINEAR [26], which supports the one-vs-all strategy for multi-class classification. In binary SVM classification for a given set of instance labels $(x_i, y_i)$ where $i = 1,..,l$, $x_i \in R^n$, $y_i \in \{-1, +1\}$, the function $\max(1-y_i w^T x_i, 0)^2$ is referred to as L2-SVM and is the most appropriate common loss function for the linear SVM classifier in LIBLINEAR. For the multi-class case, LIBLINEAR implements a method by Crammer and Singer.

Our dataset for training the coarse level SVM classifier (henceforth referred to as SOC-Classifier) was a set of 3.6m jobs (each composed of titles, descriptions and requirements) listed on Careerbuilder.com in 4Q 2012. It is a daunting task to obtain a labeled set of training data for large datasets. While crowdsourcing is an option, it is expensive and there are likely to be systematic and consistency errors [20]. We were able to circumvent this by tagging our jobs dataset using a $3^{rd}$-party O*NET classification system. One issue that we encountered was a class imbalance problem where classes such as SOC-55 (military jobs) had little representation in the dataset compared to jobs from the SOC-15 category (engineering jobs). For the time being we resolved this by under-sampling all the classes to a base count of 150k per class which allowed training data for almost all the SOC categories to be included in the training set. This reduced the training dataset down to 1.8m jobs but it was still larger than what the existing system was trained on. We chose features based on frequency and removed any feature that occurred in fewer than two training instances. The total number of features was 10m. The SOC-Classifier was trained on an m2.2xlarge EC2 instance with 34.20GB RAM and 13 compute units in under 1.5 hours. The resulting classifier exhibits very fast classification performance. Single-node classification on instances is extremely fast at 60ms or less. On a 40-node cluster, the Hadoop[6]-based, distributed, batch SOC-Classifier system can classify a 4 million jobs dataset within a few minutes. For a multi-class text classifier such as the SOC-Classifier, the appropriate performance metrics are average precision, average recall, and macro-average $F_1$ score [27]. We ran a k-fold cross validation test with k set to 10 to estimate the accuracy of the model generated. The general consensus in the machine learning community is that 10-fold validation provides the best balance between overlap between test sets and the size of test sets [28]. We achieved an average precision of 97.49%, average recall of 94.80% and an average $F_1$ score of 96.11%. Coverage was very high at 99%. Since one of our goals is to build an automated O*NET code classifier at least comparable to the $3^{rd}$-party system, we compared the results of the SVM classifier with this system on a jobs dataset of 2.1m from 1Q 2013. With the $3^{rd}$-party system as the gold standard, the SOC-classifier achieved an average classification accuracy of 89.92% for the top-level two-digit SOC code.

We created vertical classifiers for healthcare (a combination of SOC-29-Healthcare practitioners and SOC-31-Healthcare support) and SOC-15-computer and mathematical occupations categories. Using this two-stage approach resulted in more niche titles being discovered because it enabled the computationally intensive clustering step to operate entirely on datasets for a particular category. For example the SOC-15 vertical contained titles such as *Cloud Engineer* and *Salesforce Developer* that do not exist in the current system. For the healthcare vertical classifier we also conducted an A/B test [1] on the CareerBuilder.com website using live traffic to compare the performance with the existing classifier. Our main metric for this comparison was the conversion rate of users which is defined as Expression Of Interest (EOI)/Unique Visitors (UV). We register an EOI when a visitor clicks on the *Apply Now* button for a job. We ran the A/B test for three weeks and experienced a 1.7% gain in the conversion rate. We are aware that our vertical classifier is a multi-label classifier and we will conduct further evaluation of its performance using metrics such as Hamming loss and zero-one loss.

---

[5] http://lucene.apache.org

[6] http://hadoop.apache.org

## 4. CONCLUSION AND FUTURE WORK

In this paper we described the existing semi-supervised, proximity-based, multi-class multi-label job title classification system currently in operation at CareerBuilder.com. We discussed the limitations of the system and proposed an enhancement in the form of a coarse-level and fine-level cascading architecture composed of an SVM classifier and a k-NN classifier. This enables us to make progress towards our goal of having an automated O*NET job title taxonomy classification system as well as enhancing our parallel pipeline of more fine grained job title labels. Early A/B tests indicate that there is potential for increased visitor engagement on the website due to better, more accurate and detailed classifications of jobs. Our future work involves building the pipeline for the fine-level classification at the 4-digit O*NET-SOC code, creating more vertical classifiers for the job title classifier pipeline, conducting A/B tests for the vertical classifiers, and eventually releasing the job classification API for external consumption.

## 5. ACKNOWLEDGMENTS
We would like to thank Danthanh Tran for helping with A/B tests and the classification API.